# Ending-based Strategies for Part-of-speech Tagging

Greg Adams, Beth Millar, Eric Neufeld and Tim Philip
Department of Computational Science
University of Saskatchewan
Saskatoon, SK, Canada, S7N 0W0
eric@spr.usask.ca

## Abstract

Probabilistic approaches to part-of-speech tagging rely primarily on whole-word statistics about word/tag combinations as well as contextual information. But experience shows about 4 per cent of tokens encountered in test sets are unknown even when the training set is as large as a million words. Unseen words are tagged using secondary strategies that exploit word features such as endings, capitalizations and punctuation marks.

In this work, word-ending statistics are primary and whole-word statistics are secondary. First, a tagger was trained and tested on word endings only. Subsequent experiments added back whole-word statistics for the $N$ words occurring most frequently in the training set. As $N$ grew larger, performance was expected to improve, in the limit performing the same as word-based taggers. Surprisingly, the ending-based tagger initially performed nearly as well as the word-based tagger; in the best case, its performance significantly exceeded that of the word-based tagger. Lastly, and unexpectedly, an effect of negative returns was observed — as $N$ grew larger, performance generally improved and then declined. By varying factors such as ending length and tag-list strategy, we achieved a success rate of 97.5 percent.

## 1 INTRODUCTION

The dominant approach in natural-language processing (NLP) is a knowledge- and inference-based cognitive approach. Although probabilistic approaches to linguistic problems were attempted earlier in the century (Zipf, 1932), they were hampered by the very real difficulties of collecting meaningful statistics and of performing subsequent calculations. Recently probabilistic approaches have overcome these difficulties with the availability of electronic corpora such as the LOB Corpus (Johansson, 1980; Johansson et al., 1986), the Brown Corpus (Kučera and Francis, 1967) or the UPenn corpus (Santorini, 1990), as well as the existence of powerful and inexpensive computers.

The approaches appear to trade accuracy for generality. The traditional knowledge-based approach is highly accurate and sensitive within a narrow domain of discourse, but prone to catastrophic failure outside its limited domain. In contrast, the statistical or corpus-based approach is robust because it can deal with a wide range of situations. By its nature, however, the statistical approach admits some error in most situations because statistical summaries sometimes wash distinctions out of data. Thus, the flexibility of the statistical approach makes it an attractive choice for applications, but it remains a challenge to bring error rates to levels practical for applications.

One success of the probabilistic approach has been using hidden Markov models (HMMs) for attaching part-of-speech (POS) tags to unrestricted text, often considered to be a first step towards more difficult tasks such as parsing, text-to-speech applications, grammar/style checkers, OCR and machine translation. In the course of trying to minimize the error rate in a part-of-speech tagger, we found an unexpected result, namely that too much knowledge (in the sense of too many statistical parameters) can be a bad thing. We began by training and testing the tagger on three letter endings, and in subsequent experiments, adding back whole-word statistics for the $N$ most frequently occurring words in the training set. We found that the success rate increased with $N$ up to a point and then declined. Many factors were varied, and in the best case we achieved a success rate of 97.5 per cent, the highest rate we have seen reported for a tagger that tags unseen or unknown words without any benefit of external knowledge.

The rest of the paper is organized as follows. In the next section we describe HMMs as applied to POS tagging and related terminology and notation. In Section 3, we describe our tagger and what parameters were varied in the experiments. Finally, we discuss the meaning of the results and their significance.

## 2   BACKGROUND

Much work on probabilistic POS tagging uses an electronic corpus such as the tagged LOB Corpus (Johansson *et al.*, 1986), which contains 500 text samples of approximately 2000 words distributed over 15 text categories. Each word in the LOB is accompanied by one of a set of about 150 possible word-class or POS tags. Typically, a POS tagger is trained on a large subset of the LOB and then tested on a smaller subset which can be checked for tagging accuracy.

A common approach to POS tagging is the *hidden Markov model* (HMM) (Jelinek, 1985; Church, 1989; Foster, 1991; Merialdo, 1990; Kuhn and De Mori, 1990) or variations thereof (DeRose, 1988; Garside *et al.*, 1987) where language is assumed to be produced by a hidden model that cannot be observed directly but whose effects can be observed. For a good introduction and overview, see (Charniak *et al.*, 1993). Given a sequence of tokens $w_1 \ldots w_n$, an actual stream of text of length $n$ we abbreviate to $\wedge_{i=1}^{n} w_i$, the HMM method computes the word-tag sequence (or simply tag sequence) $t_1 \ldots t_n$ (abbreviated $\wedge_{i=1}^{n} t_i$) that most probably generated the sequence. The sequence may be an entire sentence or, in the case of a bi-tagger (defined below), a sequence of tokens beginning and ending with an *unambiguous* token, a token with only one known tag assignment. Because language is inherently ambiguous, generally there are many reasonable possibilities, even in context. A famous example is "time flies", but it is easy to produce others. Thus, the HMM-based approach chooses the tag sequence that maximizes

$$P(\wedge_{i=1}^{n} t_i \mid \wedge_{i=1}^{n} w_i). \qquad (1)$$

By the product rule of probability,

$$P(\wedge_{i=1}^{n} t_i, \wedge_{i=1}^{n} w_i) = P(\wedge_{i=1}^{n} t_i \mid \wedge_{i=1}^{n} w_i) \cdot P(\wedge_{i=1}^{n} w_i), \qquad (2)$$

and since the last term of the right hand side is invariant over all tag sequences, the problem is equivalent to maximizing the left hand side.

Two assumptions are commonly made in probabilistic POS tagging: 1) that the probability of any tag $t_i$ directly depends only on the $k$ tags immediately preceding it, and 2), that the probability of any word $w_i$ depends only upon the tag that produced it. A sequence of $k$ tags is called a $k$-gram. When $k = 1$, the tagger is called a *bi-tagger* and a sequence of $k + 1$ tags is called a *bigram*; for $k = 2$, the corresponding terms are *tri-tagger* and *trigram*. Since performance improves only marginally for $k > 1$ (Foster, 1991), we use $k = 1$. Using the product rule of probability as well as these independence assumptions, it is easy to show Equation 1 (or equivalently, the left hand side of Equation 2) is maximized when

$$\prod_{i=1}^{n+1} P(w_i \mid t_i) P(t_i \mid t_{i-1}) \qquad (3)$$

is maximized, where $n$ is the number of tokens in the sequence, and $t_0, t_{n+1}, w_{n+1}$ denote dummy word-tags and words at the beginning and end of the sequence.

It is well known that a maximum value for this simplified expression can be computed in time linear in $n$. The probabilities in Equation 3 parameterize the HMM and are easily estimated from tagged electronic corpora such as the LOB.

An interesting problem is handling unseen words (Adams and Neufeld, 1993; Church, 1989; Foster, 1991; Merialdo, 1990; Meteer *et al.*, 1991; Kupiec, 1992), that is, words not occurring in the training corpus, and therefore words for which probabilities are not known. Testing the tagger on a subset of the training corpus or only on known words (Foster, 1991; Meteer *et al.*, 1991) inflates accuracy because much of the vocabulary is used infrequently. About 50 per cent of the words in the LOB appear exactly once, so it is not surprising that many words are encountered for the first time in the test corpus. For example, (Adams and Neufeld, 1993) reports that after training a tagger on a 900,000 token subset of the LOB corpus, about 3700 of 100,000 tokens in the test corpus are unseen. Most of these unseen words (for example, *outjumped* and *galaxy*) in our view are neither exotic nor highly specialized but simply reflect the vastness of human experience.

Many HMM-based approaches to tagging text containing unseen words operate on the following principle — lexical probabilities computed on whole-word statistics form the primary strategy of the HMM, and secondary strategies are used for unseen words. As an example of such a secondary strategy, the tagger may assign equal probabilities to word/tag combinations and let contextual probabilities "do the work" (Meteer *et al.*, 1991). In other work, Church (1989) uses capitalization to identify unseen proper nouns. Meteer *et al.* (1991) attack unseen words by identifying combinations of prominent word features, in particular, statistics on a definitive set of 32 predefined inflectional and derivational word endings, such as *-ed* and *-ion*. Building on this approach, and observing that such a set must be chosen by a native speaker of the language, in (Adams and Neufeld, 1993) statistics are collected on arbitrary 2-, 3- and 4-letter word endings and other prominent word features such as capitalization or punctuation. That work also uses an external lexicon with no statistics but containing associated part of speech tags with words that helps the tagger to avoid bad guesses. Kupiec (1992) uses estimates of frequency of features with tags. Other approaches might include extraction of roots. Various combinations of these secondary strategies give success rates on unseen words alone from low (43.9 per cent) to reasonable (85.2 per cent) with overall success rates as high as 97 per cent; all techniques appear to improve the performance of the tagger to some extent.

-ne -of -he -at -es -of -he -ic -ch -as -en -in -he -in -of -ch -ng, -he -ss -of -ng -ts -of -ch -to -ed -xt. -is -is -ly -ed -as -a -st -ep -ds -re -lt -ks -ch -as -ng, -ch -es -ic -ge.

(a) Sample text, all words truncated to two letters.

-ne of the -at -es of the -ic -ch has been in the -in of -ch -ng, the -ss of -ng -ts of -ch to -ed -xt. This is -ly -ed as a -st -ep -ds -re -lt -ks such as -ng, which -es -ic -ge.

(b) Same text, truncated plus closed class words

One of the great -es of the -ic approach has been in the -in of -ch -ng, the process of -ng parts of speech to -ed -xt. This is generally considered as a first -ep towards more difficult -ks such as -ng, which -es -ic knowledge.

(c) Same text, 1000 most frequent words added.

One of the great -es of the -ic approach has been in the -in of -ch -ng, the process of -ng parts of speech to -ed text. This is generally considered as a first step towards more difficult -ks such as -ng, which -es -ic knowledge.

(d) Same text, 3000 most frequent words added.

Figure 1: A little knowledge helps a lot.

## 3  THE EXPERIMENTS

This work attempts to take a purely syntactic approach. Figure 1 motivates the idea. Word endings are the primary strategy and whole-word statistics are secondary. Because the idea was to see how successful a tagger could be by training on frequency information alone, no attempt was made to extract roots or use external lexicons. The tagger was initially trained and tested using statistics for fixed-length word endings alone. In one set of experiments, an exception was made for so-called "closed-class" words, words that belong to classes whose membership is not expected to change with time — pronouns, conjunctions, state of being verbs, and so on. Using this approach, Equation 3 is replaced by

$$\prod_{i=1}^{n+1} P(e_i|t_i)P(t_i|t_{i-1}) \qquad (4)$$

where $e$ denotes word ending.

In later experiments, whole-word statistics for the $N$ most frequently occurring words were added to the model, based on the assumption that the most frequent words are the oldest and most irregular words.

Subsequent experiments varied two other parameters — the technique for estimating the probability of unseen tag sequences and the effective tag-list strategy. Effective tag-list strategies are defined and discussed below; as regards estimating unseen tag sequences, even with a 900,000 word training set, many possible tag sequences don't occur and the question arises how to estimate them. The zero (maximum-likelihood) estimate seems reasonable — sometimes it is better in the long run to use what you know rather than guess at possibilities. However, zero probabilities may exclude reasonable tag-sequences in favour of other extremely unlikely but non-zero sequences. A standard solution to this problem is the so-called "add-one" strategy used in engineering where an observed frequency $r$ is adjusted by

$$r^* = r + 1.$$

Observing that the "add-one" strategy may overcompensate infrequently occurring bigrams, Church and Gale (Church and Gale, 1991) find the Good-Turing estimator,

$$r^* = \frac{(r+1)N_{r+1}}{N_r},$$

where $N_r$ is the number of bigram sequences occurring $r$ times, outperforms both maximum likelihood and "add-one". We also compared the "add-one" estimate against the Good-Turing estimator (Good, 1953).

For any word, its *effective tag-list* (ETL) is the set of all tags occurring with that word in the training corpus. The ETL strategies can be approximately characterized as estimating word/tag probabilities by using simple word-ending statistics versus using word-ending statistics normalized over the set of whole-word possibilities. For example, the token *precise* occurs relatively few times in the LOB but only with tag type JJ. The token *these* occurs only with tag type DTS in the training set. Yet if both words truncated retain their word-based unit ETLs, the truncated tokens can be treated unambiguously by the tagger. It is reasonable to ask whether the range of possibilities might be de-fined by the whole-word information yet estimate the probabilities from the endings.

Table 1: Tokens Correctly Tagged

| Experiment | \multicolumn{7}{c}{Number of Most Frequent Words Put Back In LOB} | | | | | | |
|---|---|---|---|---|---|---|---|
| | 0 | 1,000 | 5,000 | 10,000 | 20,000 | 30,000 | 45,000 |
| **2 letter endings** | | | | | | | |
| Unit ETL | 89.9 | 96.3 | 97.1 | 97.2 | 97.3 | 97.2 | 96.1 |
| Unit ETL + GT | 89.9 | 96.3 | 97.2 | 97.3 | 97.4 | 97.5 | 96.5 |
| Relexed | 83.9 | 93.7 | 95.6 | 96.1 | 96.3 | 96.5 | 95.3 |
| Relexed + GT | 83.9 | 93.7 | 95.6 | 96.1 | 96.5 | 96.8 | 95.8 |
| **3 letter endings** | | | | | | | |
| Unit ETL | 94.4 | 96.5 | 97.0 | 97.1 | 97.0 | 96.9 | 96.1 |
| Unit ETL + GT | 94.4 | 96.6 | 97.1 | 97.2 | 97.3 | 97.2 | 96.5 |
| Relexed | 91.2 | 94.7 | 95.9 | 96.2 | 96.3 | 96.3 | 95.5 |
| Relexed + GT | 91.2 | 94.7 | 95.9 | 96.3 | 96.6 | 96.7 | 96.2 |
| **4 letter endings** | | | | | | | |
| Unit ETL | 95.9 | 96.4 | 96.7 | 96.7 | 96.6 | 96.4 | 95.6 |
| Unit ETL + GT | 96.0 | 96.5 | 96.8 | 96.8 | 96.8 | 96.7 | 96.0 |
| Relexed | 94.6 | 95.5 | 95.9 | 96.1 | 96.1 | 96.0 | 95.3 |
| Relexed + GT | 94.7 | 95.6 | 96.1 | 96.3 | 96.4 | 96.5 | 95.8 |

Should the answer be affirmative, there are several advantages. The immediate advantage is an effective reduction in search space. Secondly, this suggests that we need not collect statistics over many millions of words to obtain accurate statistics about the distribution of tag types for unusual words. Instead, we can use electronic dictionaries that define the range of logical possibilities for words, normalized over word-ending probabilities.

For reasons related to the implementation, the strategy of estimating word/tag probabilities by the complete set of ending/tag probabilities is called the *relexed* method and the strategy of favouring unit tag-lists is called the *unit ETL* strategy.

## 4   RESULTS

The expectations were that by using an ending-based strategy as primary and a word-based strategy as secondary,

1. the ending-based strategy alone would achieve a modest success rate,
2. the success rate would increase as more whole-word statistics were added, and
3. in the limit, after adding all whole-word statistics back into the model, the performance would

be identical to the approach of using whole-word statistics as primary.

Actual results differed on all counts. Instead,

1. initial performance on word endings alone was about as good as performance on whole words alone,
2. performance improved, then degraded as whole-word information was added, but,
3. in the best case, the new strategy outperformed the old.

The first experiment tested the tagger on word-ending length $L = 3$ by truncating all tokens in the training and test corpus. The subsequent success rate of 94.4 per cent (see the first column of figures, line 6, of Table 1) compared favorably with those obtained by other taggers using only whole-word statistics (between 90 per cent and 95.4 per cent) (Meteer *et al.*, 1991; Adams and Neufeld, 1993) and guessing randomly at unseen words. Then, fixing $L = 3$, a series of experiments added back whole-word statistics for the $N$ most frequently occurring words in the training set, with the expectation that "more specific" statistics would get at distinctions washed out by ending-based statistics alone. The experiment was repeated for $N = 0, 1000, 5000, 10000, 20000, 30000, 45000$ (45000 is approximately the number of unique words in the training corpus). Let $\mathcal{F}_N$ denote the set of $N$ most frequent words in the training corpus. The formula

being maximized is

$$\prod_{i=1}^{n} F(w_i|t_i)P(t_i|t_{i-1}) \quad (5)$$

where

$$F(w_i|t_i) = \begin{cases} P(w_i|t_i) & \text{if } w_i \in \mathcal{F}_N \\ P(e_i|t_i) & \text{otherwise.} \end{cases}$$

Statistics were calculated by truncating all tokens in the training set not in $\mathcal{F}_N$. Note that the number of truncated tokens remaining decreases as $N$ increases. As $N$ increased, the performance of the tagger increased up to about 97.3 per cent (see line 6 of Table 1) and then declined. To ensure this was not an error, all experiments were independently duplicated. The duplicate experiments consistently reproduced the phenomena of rising and falling success rate, although slightly different success rates were obtained due to a different ETL strategy.

In all, we tested all combinations of seven values of $N$, three values of $L$ and four ETL strategies. All results appear in Table 1 and the best success rate of 97.5 per cent, for $N = 30,000$ and $L = 2$ using the Unit ETL + GT strategy, is highlighted. The effect of each strategy is discussed below. Note that the phenomena of rising and falling success rate appears in every line in the table.

There is an artificially large jump in the last column of figures because once all words are added back to the training corpus, there remains no pool of tokens from which ending statistics can be calculated and the tagger defaults to a strategy that treats all open-class tags as equiprobable. We therefore repeated the experiments by "doubling up" the corpus, that is, by concatenating two versions of the corpus, one consisting just of truncated tokens and one with the $N$ most frequently occurring words added back. Performance improved overall, but the same trend of rising and falling performance was observed in every line of the table but one, with the amount of decline ranging from 0.0 per cent to 0.7 per cent rather than from 0.5 per cent to 1.2 per cent as in Table 1.

Our explanation is that as $N$ gets large, the whole-word statistics added back to the tagger become increasingly inaccurate because they are based on small sample sizes. Similarly, ending-based statistics become increasingly inaccurate. At the point performance declines, most words added back occur only once or twice in the training set.

### 4.1 OPTIMAL WORD-ENDING LENGTH

Word-ending length $L$ was varied between two and four letters. Note the unusual reversal between the first and sixth columns of results in Table 1. As we compare results in column 1 (for $N = 0$), and hold ETL strategy constant, the success rate strictly increases with $L$. But when we compare corresponding entries in column 6 (for $n = 30,000$), success rate strictly decreases as $L$ increases!

In column 1, it could be argued that as $L$ increases, more specialized knowledge is added to the database, increasing the tagger's accuracy. The observations in column 6 can be explained by an argument similar to that at the end of the previous section. Because so much corpus is in whole-word form, sample sizes for word endings when $L = 4$ become small and the overall accuracy of the model declines. Thus we find ourselves dealing with a "slippery slope" argument — there are no "correct" values for $L$ and $N$; rather, performance depends on the quality of statistics being used.

### 4.2 EFFECTIVE TAG-LISTS

The first set of experiments benefitted from whole-word information by retaining the word-based *effective tag-list* (ETL) assigned to each token as a preprocessing convenience. The tagger assumed tokens with unit ETLs were unambiguous and assigned those tokens the corresponding tag. Otherwise, it considered the full range of tag possibilities for each word ending.

This appeared to give the tagger an edge. For example, re-consider tokens *precise* and *these* which occur in the training set only with tags JJ and DTS, respectively. Even after truncation to two letters, both tokens always retain a unit ETL which guarantees correct tag assignment. To study the effect of this knowledge, a separate ETL strategy was devised where the training set was "re-lexed" *after* truncation of tokens, that is, each token was given an *ending-based* ETL representing all tag assignments to that word ending.

Thus, for each $L, N$ combination, four experiments were performed:

**Unit ETL** — each truncated token was left with its original word-based ETL.

**Unit ETL + GT** — each truncated token was left with its original ETL and the Good-Turing method was used.

**Relexed** — each token assigned a new ETL after truncation.

**Relexed + GT** — as above, but also using the Good-Turing method.

Generally, Unit ETL wins over relexed corpus, suggesting that for unambiguous tokens, whole-word statistics are a winning strategy. The GT method consistently offers a marginal advantage.

### 4.3 STATISTICAL SIGNIFICANCE OF THE RESULTS

Given differences between experiments as little as 0.1 per cent, it is necessary to ask whether they can be explained by the margin of error alone, and, if not, whether achieving such improvements are of practical value.

With respect to practicality, note that it seems easy to produce taggers with 95 per cent success rates. This means one error every 20 words, which is unacceptable when one considers that errors multiply as strategies combine when the tagger is used in applications such as grammar checking. It seems reasonable to suggest a tagging rate of 99 per cent or better is required for practical applications. If so, any consistent reduction of the 5 per cent error rate is meaningful.

As regards significance, the question can be framed as a test of significance of the difference between two proportions, where the null hypothesis is that the improvement (degradation) in performance is due to chance. The usual calculations show that when the difference is 0.1 per cent (very small improvement), we reject the null hypothesis at the 20 per cent significance level. When the difference is 0.2 per cent, we reject the null hypothesis at the 0.5 per cent significance level and for a difference of 0.5 per cent, we reject the null hypothesis at the 0.1 per cent significance level.

Thus, the corpus size is large enough to give confidence in changes greater than 0.1 per cent. In particular, the highest success rate of 97.5 per cent achieved for ending-based approaches as compared to success rates of 97.0 per cent for word-based approaches cannot be attributed to chance.

## 5   DISCUSSION AND FUTURE WORK

This work makes several contributions. Firstly, it attempts to take as purely syntactic an approach to POS tagging as possible. The only knowledge incorporated in the tagger is statistical knowledge about word and ending frequencies and context learned from a training corpus. No other language-specific or domain knowledge was used. However, some remarks are in order. Our focus on syntactic properties of language should not be construed as denying the value of incorporating deep knowledge into natural language processing systems or as challenging the view, widely held in the NLP community, of the importance of deep knowledge; rather it should be seen as simply testing the limits of a "purely" syntactic approach based on statistical experience. As well, it may be argued that using word endings incorporates some morphological, rather than just syntactic, knowledge into the parser. Although the endings are trivially determined by counting from the end of the word, and not by extracting roots, our results may well have been very different had we collected statistics on prefixes or even "middles".

Secondly, it seems to be the first work to consider mixed strategies of ending statistics and whole-word statistics where whole-word knowledge is sometimes discarded, and gave a best-case success rate of 97.5 per cent. We also saw that in some cases too many parameters seem to degrade performance. This suggests that a small lexicon based on good statistics is better than a huge lexicon based on poor statistics.

Perhaps of greater practical interest are the implications the results have as regards the tradeoff between the number of parameters in the HMM and the success rate, if one is interested in constructing a minimal lexicon tagger with reasonable performance in domains such as grammar checking. The marginal value of, say, doubling the lexicon is not great. This work also supports the kind of observations in (Meteer et al., 1991) that taggers perhaps don't require the huge training sets originally conjectured. In (Meteer et al., 1991), it is noted that a tagger trained on 64,000 words rather than 1,000,000 suffers a relatively small decline in performance when tested on known words only. In a similar vein, we found an ending-based tagger trained on a relexed corpus of 100,000 2-letter endings tagged a 300,000 token corpus with a success rate of 83.3 per cent (as compared to 83.9 per cent when trained on 900,000 words). To put it another way, the gains using the inexpensive techniques described here compare well with the gains achieved by increasing the training data tenfold.

Taken together, these results suggest that at least at the level of POS tagging, natural language processing may avoid at least some of the apparent need for huge lexicons, massive training sets and large numbers of parameters by focusing on quality rather than quantity of knowledge. There are other ways to implement mixed strategies. One is suggested by the result of (Adams and Neufeld, 1993), where it was observed that a tagger containing all whole-word statistics from the training corpus performed better with three-letter endings than with either two- or four-letter endings.

The present results suggest no single strategy improves performance independently of other strategies; in other work, we attempt to quantify the relationships with some hope of finding a principle to guide construction of taggers. One can simply choose the $L$, $N$ and ETL strategy that worked best in practice, but it would be interesting to find a unifying principle. For example, the Unit ETL experiments suggests there may be value in using available ending statistics, but normalizing over known possibilities for the word. Presently we are investigating a variety of mixed strategies, and applying them to the construction of HMM-based grammar checkers.

The Sound Probabilistic Reasoning Project at the University of Saskatchewan continues to investigate several variations on these experiments, but with a focus on purely syntactic approaches.

### Acknowledgments

This research was supported by the Natural Science and Engineering Research Council of Canada as well as the Institute for Robotics and Intelligent Systems. The first author thanks the City of Saskatoon for permitting educational leave, the University of Saskatchewan for a graduate scholarship, and IRIS for subsequent support. George Foster made software available. Thanks to David Haugen who gave an earlier version of this manuscript a careful reading, and thanks to the referees for their comments.